\documentclass[lettersize,journal]{IEEEtran}
\usepackage{amsmath,amsfonts}
\usepackage{algorithmic}
\usepackage{algorithm}
\usepackage{array}
\usepackage[caption=false,font=normalsize,labelfont=sf,textfont=sf]{subfig}
\usepackage{textcomp}
\usepackage{stfloats}
\usepackage{url}
\usepackage{verbatim}
\usepackage{graphicx}
\usepackage{cite}
\usepackage{xcolor}
\usepackage{color}
\usepackage{setspace}
\usepackage[symbol]{footmisc}
\usepackage{makecell}
\usepackage{graphicx}
\usepackage{amssymb}
\usepackage{pifont}
\newcommand{\cmark}{\ding{51}}%
\usepackage{hyperref}

\begin{document}

\title{Programmable-Room:\\Interactive Textured 3D Room Meshes Generation Empowered by Large Language Models}

\author{
Jihyun Kim\footnotemark*, Junho Park\footnotemark*, Kyeongbo Kong\footnotemark*, and Suk-Ju Kang
\thanks{
This research was supported by Samsung Electronics (IO201218-08232-01), the MSIT(Ministry of Science and ICT), Korea, under the ITRC (Information Technology Research Center) support program (IITP-2025-RS-2023-00260091) and the Graduate School of Metaverse Convergence support program (IITP-RS-2022-00156318) supervised by the IITP(Institute for Information Communications Technology Planning Evaluation), and the National Research Foundation of Korea(NRF) grant funded by the Korea government(MSIT) (RS-2024-00414230).
\\ \indent Jihyun Kim, Junho Park, and Suk-ju Kang are with the Department of Electronic Engineering, Sogang University, Seoul, South Korea (e-mail: jhkim5950@sogang.ac.kr; junho18.park@gmail.com; sjkang@sogang.ac.kr).
\\ \indent Kyeongbo Kong is in the Department of Electrical and Electronics Engineering, Pusan National University, Pusan, South Korea (e-mail: kkb4723@gmail.com).
\\ \indent The symbol * implies Jihyun Kim, Junho Park, and Kyeongbo Kong are equally contributed to this work. (Corresponding author: Suk-Ju Kang.)
}
}

\maketitle

\begin{figure*}[!tb]
\centering
\includegraphics[width=\textwidth]{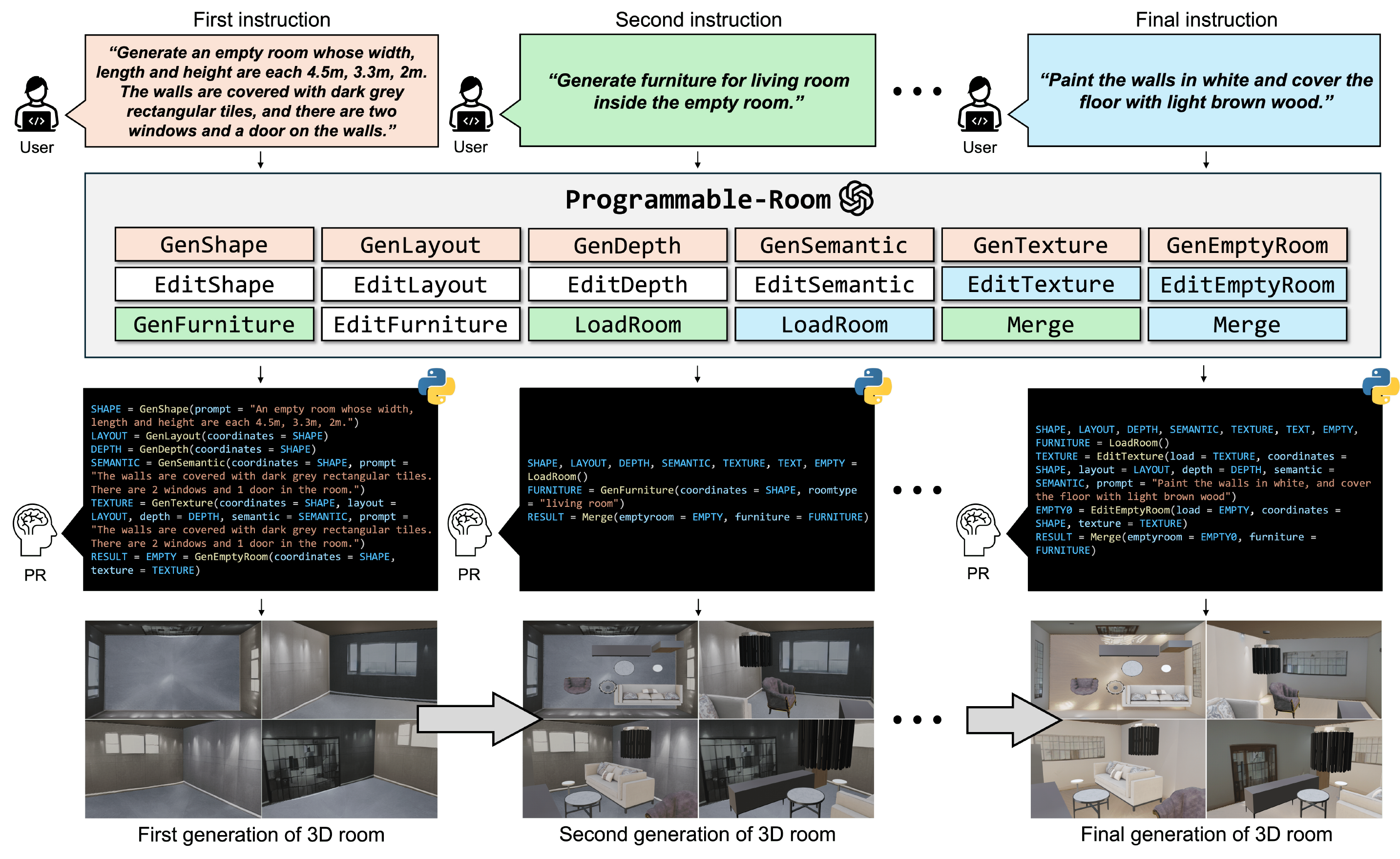}
\caption{\textbf{Overall pipeline of Programmable-Room.} When a natural language instruction is given by users, Programmable-Room (PR) converts it to a python-like visual program. Then for each line, modules supported by PR is activated. After the initial stage, users can continue editing the room mesh from the initial stage until they obtain the most satisfying result. The red, green, and blue module boxes are selected from the red, green, and blue instruction boxes, respectively.}
\label{fig:teaser}
\end{figure*}

\begin{abstract}
We present Programmable-Room, a framework which interactively generates and edits a 3D room mesh, given natural language instructions. For precise control of a room's each attribute, we decompose the challenging task into simpler steps such as creating plausible 3D coordinates for room meshes, generating panorama images for the texture, constructing 3D meshes by integrating the coordinates and panorama texture images, and arranging furniture. To support the various decomposed tasks with a unified framework, we incorporate visual programming (VP). VP is a method that utilizes a large language model (LLM) to write a Python-like program which is an ordered list of necessary modules for the various tasks given in natural language. We develop most of the modules. Especially, for the texture generating module, we utilize a pretrained large-scale diffusion model to generate panorama images conditioned on text and visual prompts (\textit{i.e.}, layout, depth, and semantic map) simultaneously. Specifically, we enhance the panorama image generation quality by optimizing the training objective with a 1D representation of a panorama scene obtained from bidirectional LSTM. We demonstrate Programmable-Room's flexibility in generating and editing 3D room meshes, and prove our framework's superiority to an existing model quantitatively and qualitatively. Project page is available in \textcolor{magenta}{\href{https://jihyun0510.github.io/Programmable_Room_Page/}{https://jihyun0510.github.io/Programmable\_Room\_Page/}}.

\end{abstract}


\section{Introduction}

\IEEEPARstart{I}{magine} creating a room of your taste. The first step would be to define the overall design, starting with the room's shape. Then, you would want to add details such as the style of the floor, walls, and ceiling. Once the basic construction is complete, the next consideration might be filling in the room with appropriate furniture. Not to mention, editing the scene numerous times, as the perfectly satisfying room is often a result of refining the details over time.
To achieve such interactive generation and modification of various components of 3D indoor scenes using language instructions, we introduce Programmable-Room, a novel approach that incorporates the visual programming (VP) \cite{gupta2023visual}.

Programmable-Room decomposes the process of building a room into subtasks, such as (1) determining 3D coordinates of a room which align with user-provided instructions, (2) generating a panorama texture image, (3) constructing an empty room by integrating the coordinates and the panorama texture image, and (4) arranging appropriate furniture in the room. The major benefit of distinct generation of various components of an indoor scene is precise control over each attribute. For example, in Fig. \ref{fig:teaser}, users can change the color and texture of the floor and walls individually without affecting other elements such as the room's shape, furniture, and the location of windows and door.

We incorporate VP to Programmable-Room to support the various decomposed tasks with a unified framework. VP is a method that leverages a large language model (LLM) to write a python-like modular program which is a list of subtasks, for a complex vision task given in natural language. Then, each line of the program is executed sequentially so that the outputs of earlier lines become the inputs of the following lines. As such, the output of the last line represents the desired result. 
Similarly, Programmable-Room employs GPT-4 \cite{OpenAI2023GPT4TR} to select a combination of predefined modules and arrange them into logically ordered Python codes for various instructions. The instructions range from increasing the width of a room to creating a fully textured and furnished room in a single step. 
Moreover, as shown in Fig. \ref{fig:teaser}, our framework allows infinite editing of the room by calling upon the necessary data in string, number, or image formats from the previous stages.

Similarly to our paper, several works \cite{hollein2023text2room, zhang2024text2nerf, fang2023ctrl, yang2024holodeck, fridman2024scenescape} aim to generate a textured 3D model of an indoor scene from text prompts. For example, Text2Room \cite{hollein2023text2room}, Text2Nerf \cite{zhang2024text2nerf} and Scenescape \cite{fridman2024scenescape} generate 3D indoor scenes by progressively updating a 3D model frame by frame using images from various viewpoints. Ctrl-Room \cite{fang2023ctrl} leverages a panorama image with a plausible room layout, and then conducts panorama depth estimation for room mesh reconstruction. However, in contrast to Programmable-Room, these models are highly limited in interactively generating and editing 3D indoor scenes.

For the convenience of users, we develop various algorithms and models in the module list: \textit{GenShape, GenSemantic, GenTexture, GenEmptyRoom, GenFurniture, EditShape, EditLayout, EditDepth, EditSemantic, EditTexture, EditEmptyRoom, EditFurniture, LoadRoom,} and \textit{Merge}. Especially for \textit{GenFurniture} which aims to generate room texture images in accordance with both the texture and the shape specified by users, we introduce a generative model, panorama room image generation (PRIG). It is a diffusion\cite{rombach2022high}-based model that generates a panorama room texture image conditioned on text and visual prompts, and they convey texture and geometric information, respectively. Similar works \cite{zhang2024taming, chen2022text2light, tang2023mvdiffusion}, which aim to the panorama indoor scene image generation, support texture control only; they cannot control shape or spatial information. 
PRIG is designed to utilize multiple visual prompts, \textit{i.e.}, layout, depth, and semantic maps, because a single visual prompt is insufficient for generating a high-fidelity panorama image with respect to the specified geometric information. Thus, inspired by Uni-ControlNet \cite{zhao2023uni}, we employ a multi-scale injection strategy and feature denormalization (FDN) to effectively condition the diffusion model on multiple visual prompts. Additionally, we optimize the training objective with the 1D representation of a panorama scene obtained from bidirectional LSTM (BiLSTM) \cite{cui2018deep}. Encoding the layout of the panorama room into 1D presentation using BiLSTM offers significant advantages. This approach utilizes fewer parameters to train PRIG, captures a long-range geometric pattern of room layout, and enhances PRIG's robustness to complex layouts.

One of the major benefits of our framework is that any algorithm can be used for each module if the inputs and outputs are constant. As a result, Programmable-Room features extensibility, allowing anyone to easily add new modules or update existing ones in a plug-and-play manner whenever new models are released.
Thus, Programmable-Room is expected to execute a wider range of instructions and, most importantly, produce better results over time. 

We prove superiority of our Programmable-Room with respect to generating and editing 3D room meshes. Moreover, we demonstrate PRIG’s excellence in generating panorama room texture images. Experimental results show that our framework and PRIG are outstanding compared to state-of-the-arts quantitatively and qualitatively.

In summary, our main contributions include:
\begin{itemize}

\item We introduce Programmable-Room which interactively generates and edits a 3D indoor scene given natural language instructions with precise control. We utilize LLMs to write a Python program which is an ordered list of predefined modules, for the various decomposed tasks. 

\item We present PRIG, which generates panorama images conditioned on text and visual prompts (\textit{i.e.}, layout, depth, and semantic maps) simultaneously. To accelerate the performance, we optimize the training objective with a 1D representation of a panorama scene obtained from a bidirectional LSTM.

\item Programmable-Room is highly extendable, allowing new modules to be easily added or existing ones updated in a plug-and-play manner whenever new models are released. This flexibility ensures continuous performance improvements as related technologies advance.

\end{itemize}

\section{Related Works}

\subsection{Indoor Scene Synthesis}

Indoor scene synthesis aims to create a reasonable furniture layout in 3D spaces.
Graph-based methods, i.e, SceneFormer \cite{wang2021sceneformer}, ATISS \cite{paschalidou2021atiss}, CommonScenes \cite{zhai2024commonscenes}, and DiffuScene \cite{tang2023diffuscene}, leverage nodes and relationships of the scene.
On the other hand, geometry-aware methods, \textit{i.e.}, GAUDI \cite{bautista2022gaudi}, RGBD2 \cite{lei2023rgbd2}, CC3D \cite{bahmani2023cc3d}, RoomDreamer \cite{song2023roomdreamer}, SceneScape \cite{fridman2024scenescape}, and RoomDesigner \cite{zhao2023roomdesigner}, synthesize 3D scene with geometric information. On the other hand, LEGO-Net \cite{wei2023lego} introduces a data-driven method that learns to re-arrange the position and orientation of objects in various room types, and LayoutGPT \cite{feng2024layoutgpt} implements LLMs to arrange furniture of the 3D indoor scene. Therefore, we fully exploit the advantage of aforementioned methods by factorizing the furniture arrangement model, such as LayoutGPT, into specific module of our Programmable-Room. 

\subsection{Text-based Generation}

Text has been widely used for generating images, music, and so on \cite{jeong2023virtuosotune, tan2021cross, peng2021knowledge, yuan2023semantic}. Specifically, text-to-image (T2I) generation aims to generate realistic images from texts.
In an effort to guarantee semantic consistency between the text description and generated images, extensive research has been conducted. \cite{tan2021cross} proposes a textual-visual semantic matching module of which a partial loss function is optimized for a better feature extraction. \cite{peng2021knowledge} introduces constructed knowledge base for more vivid images. \cite{yuan2023semantic} suggests a two-fold semantic distance discrimination to measure image-text semantic relevance.  

Other recent researches \cite{kim4600274controlling, zhang2023adding, mou2024t2i, park2024attentionhand} focus on enhancing user control over the generation process.
For example, ControlNet \cite{zhang2023adding} and T2I-Adapter \cite{mou2024t2i} add lightweight adapters to Stable Diffusion (SD) \cite{rombach2022high}. Specifically, by freezing the weight of SD, they allow fine-tuning with a small-scale target dataset, reducing training costs, and generating a conditioned image with a single visual prompt. Howeverm when it comes to multiple visual prompts, the fine-tuning costs and model size increase.
To address this challenge, Uni-ControlNet \cite{zhao2023uni} categorizes conditions into two groups: local and global control, achieving efficiency in terms of training costs and model size. 

In the field of panorama room image generation, the main issue is to preserve structural information of the room. Text2Light \cite{chen2022text2light} generates a high-quality HDR panorama image, and MVDiffusion \cite{tang2023mvdiffusion} produces a high-resolution image or extrapolates a perspective image to a 360-degree view. However, they can generate a panorama image conditioned only on a text prompt but not on multiple visual prompts, limiting user control. 

Recently, these have been efforts to generate a 3D scene from a text prompt. For example, Text2Room \cite{hollein2023text2room} and Text2Nerf \cite{zhang2024text2nerf} generate 3D indoor scenes by progressively updating a 3D model frame-by-frame using images from various viewpoints. However, they tend to ignore global contexts as they rely on partial and local contexts during iterative image generation. Ctrl-Room \cite{fang2023ctrl} leverages a panorama image with a plausible room layout, and then performs panorama depth estimation for mesh reconstruction. Similarly to previous work \cite{hollein2023text2room, zhang2024text2nerf}, Ctrl-Room has limitations in generating an indoor scene which is plausible when rendered from various view points. Specifically, since the depth estimator predicts only the depths of visible contents in the generated panorama image, the full geometry of objects cannot be reconstructed, resulting in final outputs that appear plausible only from certain viewpoints.

\begin{figure}[tb!]
\centering
\begin{center}
\includegraphics[width=\linewidth]{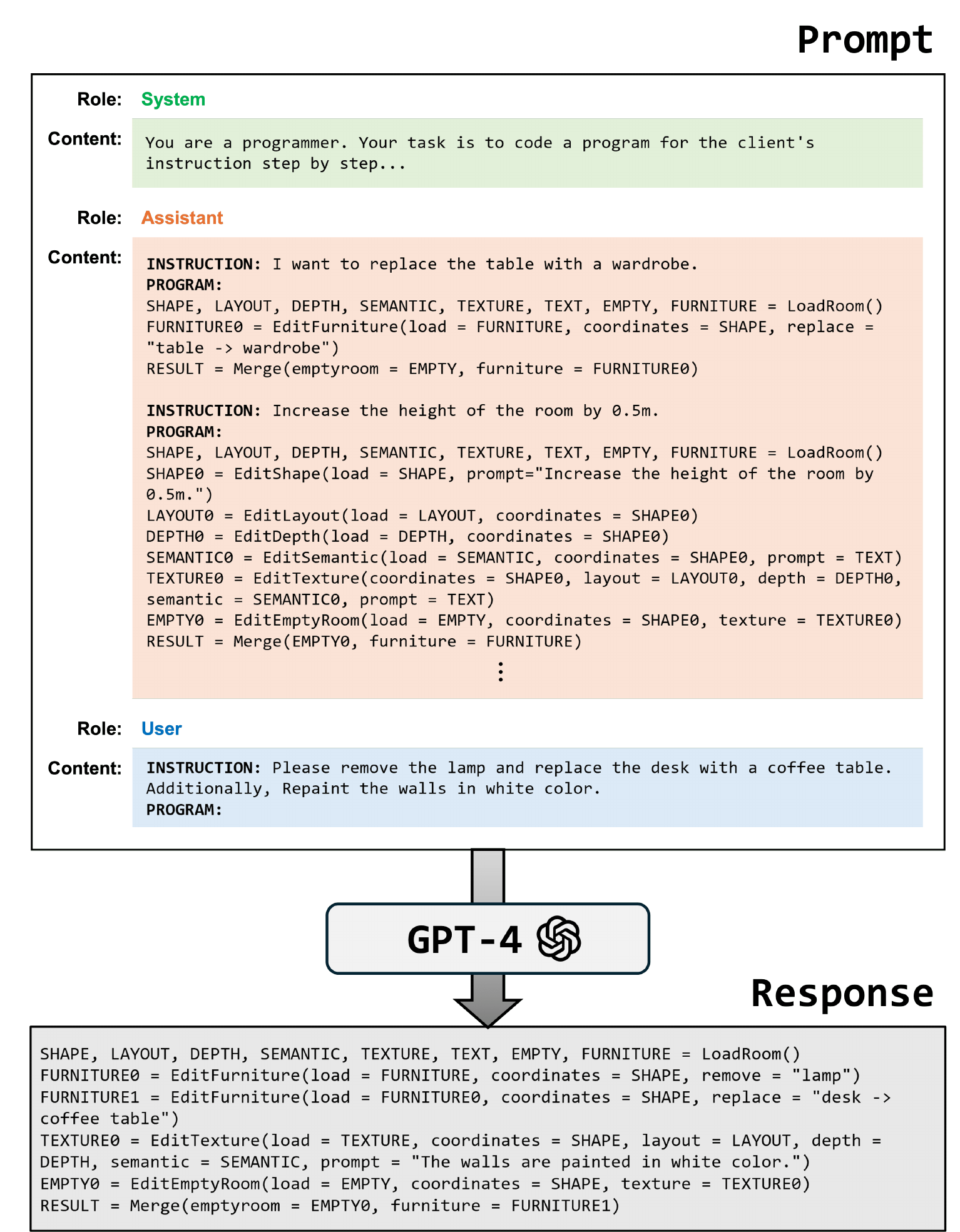}
\end{center}
\caption{\textbf{Program generation in Programmable-Room.} Given in-context examples with simple instructions, Programmable-Room infer programs for complex instructions.}
\label{fig:program_gen}
\end{figure}

\begin{figure}[t!]
\centering
\begin{center}
\includegraphics[width=\linewidth]{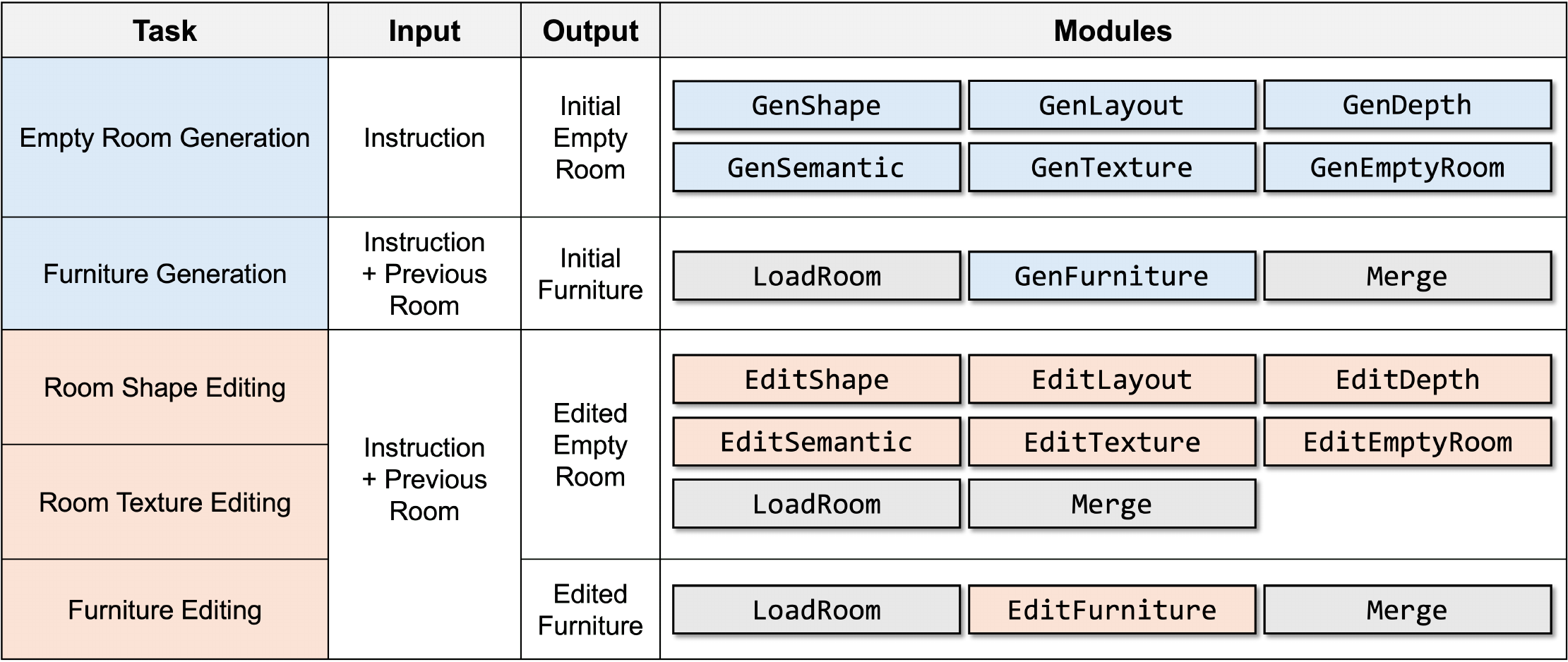}
\end{center}
\caption{\textbf{Modules supported by Programmable-Room.} Blue boxes represent modules for generating rooms and furniture, while orange boxes represent modules for editing rooms and furniture.}
\label{fig:program_modules}
\end{figure}

\section{Method}
\subsection{Overview of Programmable-Room}

Programmable-Room decomposes the challenging text-based 3D room generation task into simpler steps and performs the subtasks with specialized modules. To handle these modules, Programmable-Room employs Visual Programming (VP) \cite{gupta2023visual}, which utilizes LLMs to translate text instructions into Python-like modular programs.
To generate a certain Python program using predefined modules from natural language descriptions, LLMs need to be fine-tuned, which is not feasible due to the absence of training datasets. Therefore, Programmable-Room leverages in-context learning ability of LLMs. 

In-context learning is a way of fine-tuning LLMs to enhance their performance to certain tasks or domains, without updating the parameters. It is achieved by providing examples within the context of tasks or domains. 
Hence, as shown in Fig. \ref{fig:program_gen}, Programmable-Room utilizes GPT-4 \cite{OpenAI2023GPT4TR} with user-provided instructions and a general task description, along with pre-selected examples which consist of diverse pairs of instructions and their corresponding programs. This way, GPT-4 can act as a specialized program generator for our framework without additional training. 
Note that GPT-4 better interprets a prompt when the contents are labeled with roles. Thus, as illustrated in Fig. \ref{fig:program_gen}, we categorize an input prompt into three groups -- task descriptions into system contents; in-context examples into assistant contents; user-provided instructions into user contents. 

Moreover, as shown in Fig. \ref{fig:program_modules}, we intuitively name each module along with its input arguments and corresponding values for two primary reasons. 
Firstly, descriptive names enhance GPT-4's understanding of the purpose of each module, as well as its inputs and outputs. For instance, in Fig. \ref{fig:program_gen}, when an instruction is provided such as ``\textit{I want to replace the table with a wardrobe}", the module name \textit{EditFurniture} can be easily deduced. 
Conversely, from the name of the module \textit{EditFurniture}, which part of the sentences should be parsed and given to the module can be easily inferred.
Secondly, the resulting program is more understandable for users, enabling them to modify the in-context examples or instructions when any failure occurs. Consequently, users can construct customized LLMs that are resilient to programming errors.  

\begin{figure}[tb!]
\centering
\begin{center}
\includegraphics[width=\linewidth]{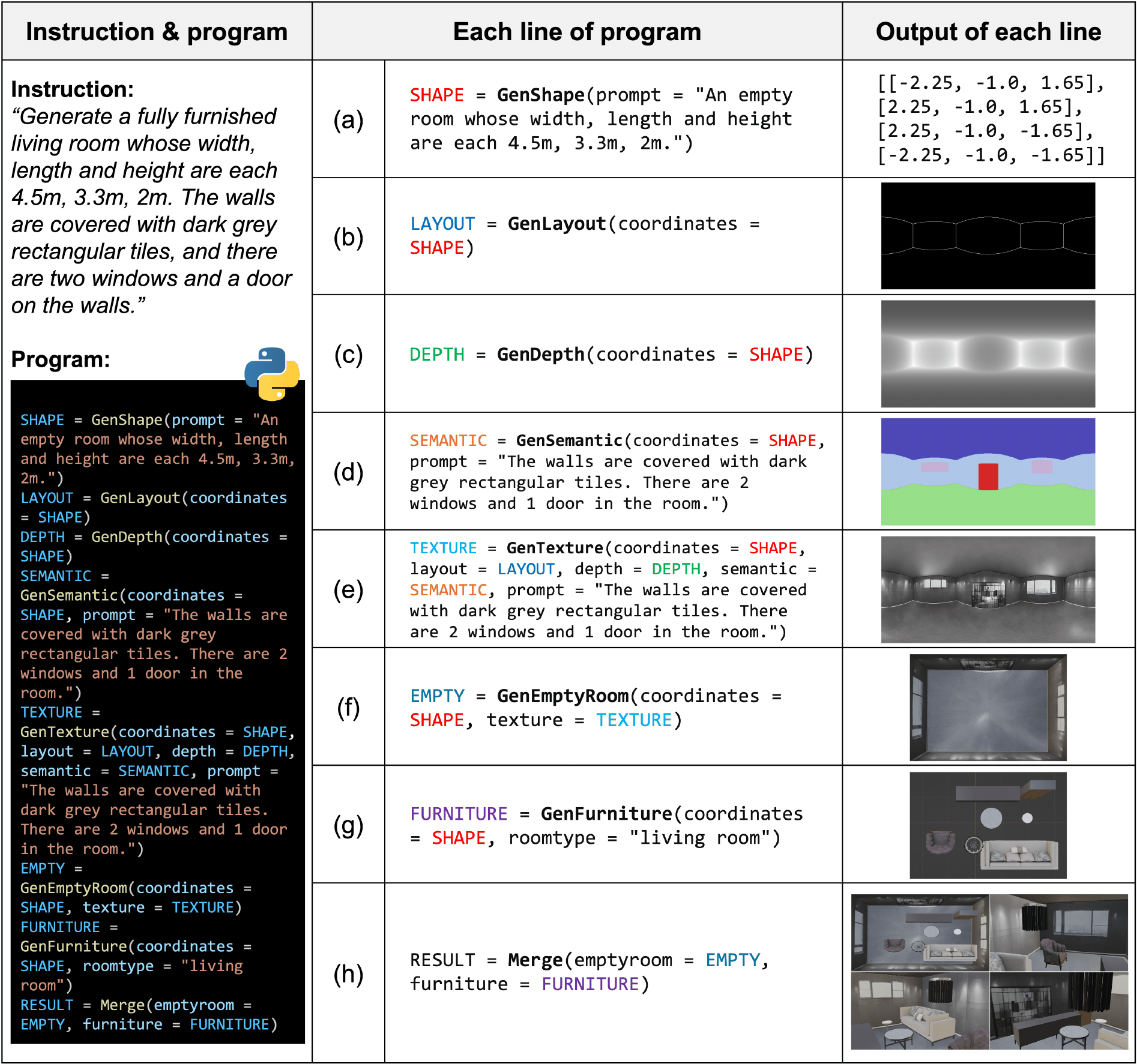}
\end{center}
\caption{\textbf{Visual outputs generated by executing each line of the program.} (a) 3D coordinates of room corners, (b-d) panorama images containing geometric information of the room, (e) textured panorama image conditioned on the room shape and texture description, (f) textured 3D room mesh, (g) generated furniture, and (h) merged room and furniture with aligned centers. For better understanding of how each output variable is used as inputs for following modules, each variables is marked in distinct colors.}
\label{fig:program_out}
\end{figure}

As illustrated in Fig. \ref{fig:program_out}, an interpreter executes the generated code sequentially, calling the relevant module with the designated inputs for each line.
The outputs are then assigned to corresponding variables, which are visualized in the right column of Fig. \ref{fig:program_out}. These variable values are subsequently inserted into subsequent modules.

\begin{figure*}[h!]
\centering
\begin{center}
\includegraphics[width=\textwidth]{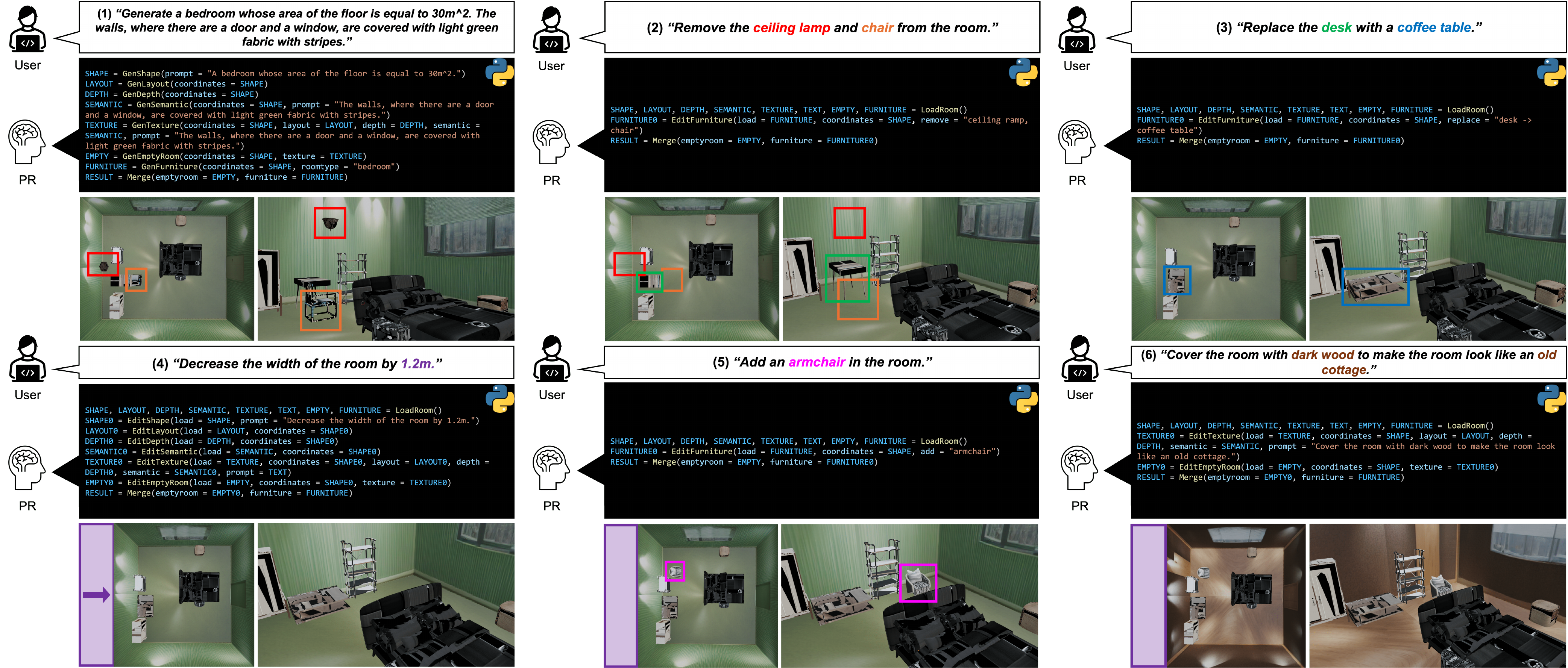}
\end{center}
\caption{\textbf{Editing examples supported by Programmable-Room.} The furniture, texture, and size of the room are edited by the text prompt given at each stage. Pairs of user instructions and corresponding bounding boxes are illustrated in the same color.}
\label{fig:editing}
\end{figure*}

\subsection{Essential Components of Programmable-Room}

Programmable-Room is the first framework to facilitate 3D indoor scene generation and editing by interlinking models and algorithms from independently studied research fields. Our technical contributions include the development of modules to address missing components in the system, encompassing 13 out of the 18 modules.

\noindent{\textbf{Room Shape Determination.}}
As shown in Fig. \ref{fig:program_out}, Programmable-Room first parses words from an instruction, which is relevant to the shape. Then, \textit{GenShape} returns a list of 3D coordinates of corners of the room. This module leverages GPT-4, which can learn visual commonsense through in-context demonstrations, to infer coordinates based on a language description. Similarly, when instructions along with the previous shape information are given, \textit{EditShape} utilizes LLMs to return the edited coordinates. Examples of editing the room's shape are illustrated in Fig. \ref{fig:editing}.

\noindent{\textbf{Textured Empty Room Generation.}}
In Programmable-Room, we suggest to generate a panorama image, then utilize its corresponding depth map to convert the image into a 3D mesh. \textit{GenEmptyRoom} is utilized for this task, where the mesh is scaled to the size specified in the instruction. In this way, we can independently control the shape and texture of the room. 

\begin{figure*}[h!]
\centering
\begin{center}
\includegraphics[width=\textwidth]{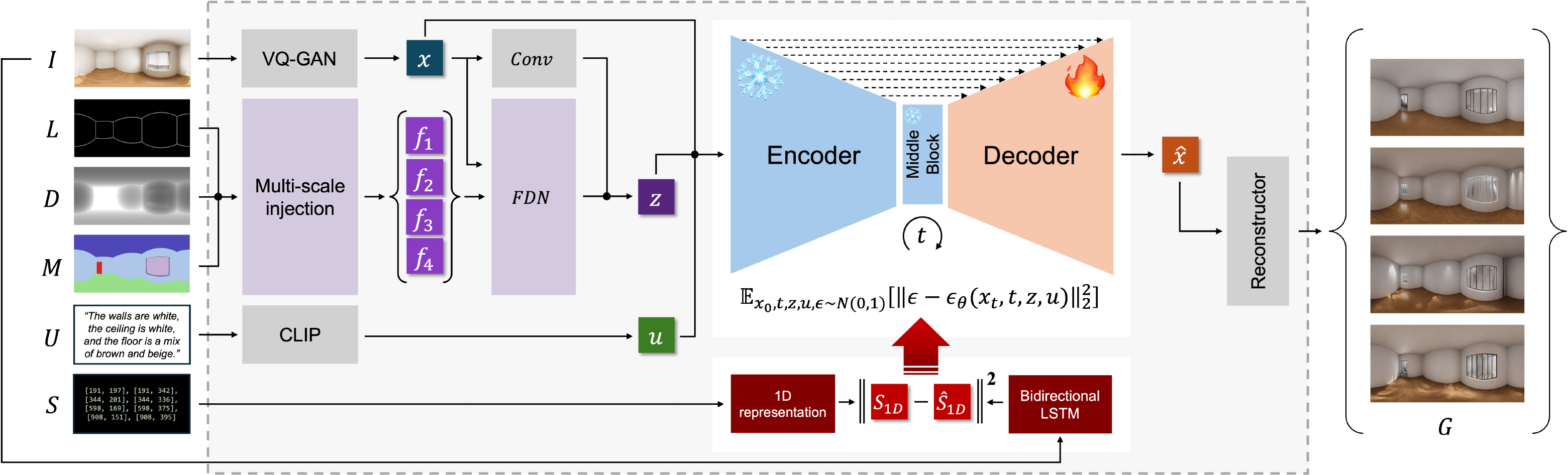}
\end{center}
\caption{\textbf{Overall pipeline of panorama room image generation (PRIG).} We utilize three latent embeddings $x$, $z$, and $u$ to train PRIG. First, $x$ is encoded from the panorama image $I$ with VQ-GAN \cite{esser2021taming}, $z$ is obtained by multiple visual prompts (\textit{i.e.}, layout map $L$, depth map $D$, and semantic map $M$) with the multi-scale injection and FDN \cite{park2019semantic} and $u$ is encoded from the text prompt $U$ with CLIP \cite{radford2021learning}. We train the U-Net-based diffusion model with three embeddings and utilize the bidirectional LSTM-based loss \cite{cui2018deep} to accelerate the performance of image generation of the feature $\hat{x}$, which is reconstructed to newly generated image $G$. The snowflakes of encoder and middle block indicate the frozen weight of Stable Diffusion \cite{rombach2022high}, and the flame of decoder indicates the learnable weight.}
\label{fig:PRIG}
\end{figure*}

For the textured panorama image generation, we develop panorama room image generation (PRIG), as research has been barely done on generating panorama image of an indoor scene based on a specific layout. 
As depicted in Fig. \ref{fig:PRIG}, the training dataset for PRIG includes: a panorama room image $I\in\mathbb{R}^{3\times1024\times512}$, a layout map $L\in\mathbb{R}^{3\times1024\times512}$, a depth map $D\in\mathbb{R}^{3\times1024\times512}$, a semantic map $M\in\mathbb{R}^{3\times1024\times512}$, a layout coordinates $S\in\mathbb{R}^{N\times2}$, and a text prompt $U$ that describes the room's texture, design, and the color. $N$ indicates the number of corners of $L$.
We take $I$, $D$, $M$, and $S$ from Structure3D \cite{zheng2020structured3d}, which is a dataset that provides rich 3D structure annotations based on panorama RGB images. Additionally, we obtained $L$ by rendering $S$, and $U$ by utilizing the off-the-shelf vision-language model \cite{bai2023qwen}. We annotated $U$ by giving a question \textit{``Describe the texture, color, and pattern of the walls, ceiling, and floor in details"}, and taking an answer such as \textit{``The walls are painted in a light blue color, the ceiling is white, and the floor is made of wood with a pattern of brown stripes."}

PRIG is a diffusion\cite{rombach2022high}-based module that generates a panorama room image from text and visual prompts, as shown in Fig. \ref{fig:PRIG}.
In general, conditional text-to-image generation models generate images conditioned on a single visual prompt. However, utilizing just a single visual prompt is insufficient for generating a high-fidelity panorama image with respect to geometric alignment. Specifically, when the generated image is solely conditioned on a layout map, the depth information of the room is significantly compromised. Conversely, conditioning PRIG with a depth map only leads to ambiguity in the boundary information of the room (\textit{i.e.}, floor-wall, ceiling-wall, and wall-wall boundary), resulting in a substantial loss of structural knowledge. Moreover, if we condition PRIG without a semantic map, categories such as ceiling, walls, floor, doors, and windows of room cannot be clearly distinguished.

Therefore, inspired by Uni-ControlNet \cite{zhao2023uni}, we design PRIG to generate conditioned images from multiple visual prompts. In other words, we utilize $L$, $D$, and $M$ as visual prompts. Subsequently, after obtaining a concatenated map $V \in \mathbb{R}^{9\times1024\times512}$ by concatenating $L$, $D$, and $M$ in channel dimension, we implement a multi-scale injection strategy. As shown in Fig. \ref{fig:PRIG}, this involves extracting features of multiple resolutions (\textit{i.e.}, $64 \times 64$, $32 \times 32$, $16 \times 16$, and $8 \times 8$) for $V$ using feature extractors $\mathcal{F}_1$ and $\mathcal{F}_2$, each comprising multiple convolution layers:
\begin{equation}
f_i = \mathcal{Z}_i(\mathcal{F}_2(\mathcal{F}_1(V))),
\end{equation}
where $f_1 \in \mathbb{R}^{192\times64\times64}$, $f_2 \in \mathbb{R}^{256\times32\times32}$, $f_3 \in \mathbb{R}^{384\times16\times16}$, and $f_4 \in \mathbb{R}^{512\times8\times8}$ indicate initial features with multiple resolutions, and $\mathcal{Z}_i$ indicates zero convolution \cite{zhang2023adding} with $i$--th resolution. 
Subsequently, we implement the feature denormalization ($FDN$) \cite{park2019semantic} to modulate normalized input noise features for $f_i$:
\begin{equation}
z_i = Conv(x) + FDN(x, f_i),
\end{equation}
where $x \in \mathbb{R}^{192\times64\times64}$ denotes the latent embedding obtained from $I$ through VQ-GAN \cite{esser2021taming}, $z_i$ denotes a conditional feature with $i$--th resolution, and $Conv$ denotes a learnable convolutional layer that allows us to obtain spatially-adaptive and learned transformation. Thus, PRIG is trained to be conditioned on two embeddings $z = \{z_i: i = 1, 2, 3, 4\}$ and $u$, which is the latent embedding of $U$ encoded by CLIP \cite{radford2021learning}.
The core benefit of using VQ-GAN for image embedding is the gain of rich feature which effectively models local and global information of an image. VQ-GAN comprises CNN, which has strength in capturing the local structure, and Transformer \cite{vaswani2017attention}, which excels at global relation modeling.

PRIG is designed based on a U-Net network \cite{ronneberger2015u}, and its encoder and middle block are initialized with the weight of a pre-trained large diffusion model \cite{rombach2022high} and fixed parameters frozen to exploit the high performance of image generation. On the other hand, parameters of the decoder are set learnable to jointly condition text and visual prompt embeddings. Similar to the state-of-the-art diffusion models, PRIG has the forward and reverse processes. In the forward process, $x_t$ is obtained by perturbing Gaussian noise on $x$ with $t$ steps. In the reverse process, a denoising network $\epsilon_{\theta}$ is trained to predict the denoised variant by denoising added noises on $x_t$. The objective for $\epsilon_{\theta}$ is as follows:
\begin{equation}
\mathcal{L}_{latent}=\mathbb{E}_{x_0,t,z,u,\epsilon \sim N(0,1)}[\|\epsilon-\epsilon_{\theta}({x_t}, t, z, u)\|^2_2],
\end{equation}
where $x_0$ indicates the initially sampled latent embedding.

Furthermore, we propose a new objective, $\mathcal{L}_{BiLSTM}$, which is a L2 loss aimed to optimize the 1D presentation obtained from a bidirectional LSTM (BiLSTM) \cite{cui2018deep}. Encoding the layout of the panorama room using BiLSTM into the 1D presentation offers significant advantages. First, the 1D representation obtained from BiLSTM not only stores the long-range layout information but also considers the left and right sides of the layout simultaneously. This effectively reflects the horizontally long panorama room layout which is continuous in both sides. Second, this 1D representation reduces the computational cost, yet comprises sufficient information as mentioned earlier. The objective is as follows:
\begin{equation}
\mathcal{L}_{BiLSTM}=||S_{1D} - \hat{S}_{1D}||^2_2.
\end{equation}
where $S_{1D}$ denotes the 1D representation of layout coordinate $S$, and $\hat{S}_{1D}$ denotes the 1D representation predicted by feeding panorama image $I$ into the BiLSTM. Therefore, the final objective is as follows:
\begin{equation}
\mathcal{L}=\lambda_{latent} \mathcal{L}_{latent} + \lambda_{BiLSTM} \mathcal{L}_{BiLSTM},
\end{equation}
where $\lambda_{latent}$ and $\lambda_{BiLSTM}$ are weighted coefficients of $\mathcal{L}_{latent}$ and $\mathcal{L}_{BiLSTM}$.
As a result, $\hat{x}$ is obtained from the decoder of PRIG, and it is reconstructed by a VQ-GAN-based decoder to generate a new panorama image $G \in \mathbb{R}^{3\times1024\times512}$. We name this decoder as a reconstructor as shown in Fig. \ref{fig:PRIG}.

During the inference stage of PRIG, we employ various conventional image processing algorithms to derive $L$, $D$, and $M$. Details on the algorithms are included in the appendix. 

\noindent{\textbf{Furniture Generation.}}
In contrast to the tasks above, 3D room layout synthesis is an active research area, which grants our framework many choices of models in allocating furniture into appropriate places. We implement LayoutGPT \cite{feng2024layoutgpt} as \textit{GenFurniture} for its cascading style sheets (CSS)-like formatting. Specifically, LayoutGPT saves furniture information such as furniture list, location, angles, and sizes in CSS file. Then, matching furniture pieces are loaded from a database into the corresponding location. For this reason, editing the furniture layout can be easily achieved by editing the CSS file, which relieves our framework from directly manipulating furniture meshes.

As shown in Fig. \ref{fig:editing}, regarding \textit{EditFurniture}, we develop methods for ``\textit{Add}", ``\textit{Replace}", and ``\textit{Remove}" functionalities using GPT-4. For adding or replacing furniture, we supply furniture layouts with room sizes similar to the current room size for in-context learning to GPT-4. The model then predicts the suitable location and orientation of the specified furniture piece in CSS format. To remove furniture, we simply delete it from the existing furniture list file.

\noindent{\textbf{Merging Furniture and Empty Room.}}
After furniture generation, Programmable-Room merges furniture with the empty room. Specifically, \textit{Merge} aligns the centers of the furniture layout and the empty room to ensure that the furniture is placed inside the room rather than at a random location.

\section{Experiment}

\subsection{Datasets}
For panorama image generation, we train and test our PRIG with Structure3D \cite{zheng2020structured3d} which is a 3D indoor scene dataset consisting of 21,773 rooms. Among the three versions (\textit{i.e.}, full, partial, and empty rooms), we utilized empty rooms for empty room generation. For each empty room, rendered images including RGB panorama, semantic panorama, depth map, and normal map are provided. Moreover, along with the images, $xy$ coordinates of intersections of walls, ceiling, and floor are available. Then, we utilized a vision-language model, Qwen-VL \cite{bai2023qwen}, to generate texts about the texture of each rendered empty scene. The dataset was split into train and test datasets with a ratio of 8:2.

For 3D room mesh generation, we utilize two datasets to furnish the textured empty room: 3D-FUTURE \cite{fu2021future} and 3D-FRONT \cite{fu2021front}. 3D-FUTURE comprises 5,000 varied scenes and involve 9,992 distinct industrial 3D CAD furniture shapes. 3D-FRONT, from which the implemented furniture generation module learns furniture arrangements, is a large-scale, and comprehensive repository of synthetic indoor scenes. It includes 18,797 rooms, each uniquely furnished with a variety of 3D objects.

\subsection{Evaluation Metrics}
For the panorama image generation, we adopt the following metrics: Fréchet Inception Distance (FID) \cite{heusel2017gans} and Kernel Inception Distance (KID) \cite{binkowski2018demystifying}.
For the 3D mesh generation, we rendered images of 10 rooms from 5 different views to conduct user study. We asked 30 participants to measure the perceptual quality (PQ) and 3D structure completeness (3DS) of the room meshes on scores ranging from 1 to 5.

\begin{table}[t!]
\centering
\caption{Quantitative comparisons on panorama image generation. PR indicates our Programmable-Room.}
\resizebox{0.7\linewidth}{!}{
\begin{tabular}{l|cc}
\Xhline{1.8pt}
{Method} & {FID$\downarrow$} & {KID$\downarrow$} \\
\hline
{Text2Light \cite{chen2022text2light}} & {103.39} & {0.15133} \\
{MVDiffusion \cite{tang2023mvdiffusion}} & {95.38} & {0.13451} \\
{PanFusion \cite{zhang2024taming}}  & {72.66} & {0.03363} \\
{PR w/o BiLSTM} & {84.89} & {0.03811} \\
{PR w/ BiLSTM (Ours)} & {\textbf{65.68}} & {\textbf{0.02354}} \\
\Xhline{1.8pt}
\end{tabular}
}
\label{tab:exp_pano}
\end{table}

\begin{table}[t!]
\centering
\caption{Quantitative comparisons on 3D mesh generation. PR indicates our Programmable-Room}
\resizebox{0.8\linewidth}{!}{
\begin{tabular}{l|cccc}
\Xhline{1.7pt}
{Method} & {PQ$\uparrow$} & {3DS$\uparrow$} & {Inference Time(s)$\downarrow$} \\
\hline
{Text2Room \cite{hollein2023text2room}} & {2.68} & {2.39} & {5179.08} \\
{Holodeock \cite{yang2024holodeck}} & {2.52} & {2.67} & {180.00} \\
{SceneScape \cite{fridman2024scenescape}} & {2.18} & {1.85} & {9300.00}\\
{PR (Ours)} & \textbf{{3.57}} & \textbf{{3.82}} & \textbf{154.61} \\
\Xhline{1.7pt}
\end{tabular}
}
\label{tab:exp_mesh}
\end{table}

\subsection{Implementation Details}
To train PRIG, we set the batch size to 3, learning rate to 1e-5, and optimizer to Adam \cite{kingma2014adam}. Both for the training and the inference stages, we set diffusion time steps to 1,000. PRIG requires a single NVIDIA A100 GPU for implementation.

\begin{figure}[t!]
\centering
\includegraphics[width=\linewidth]{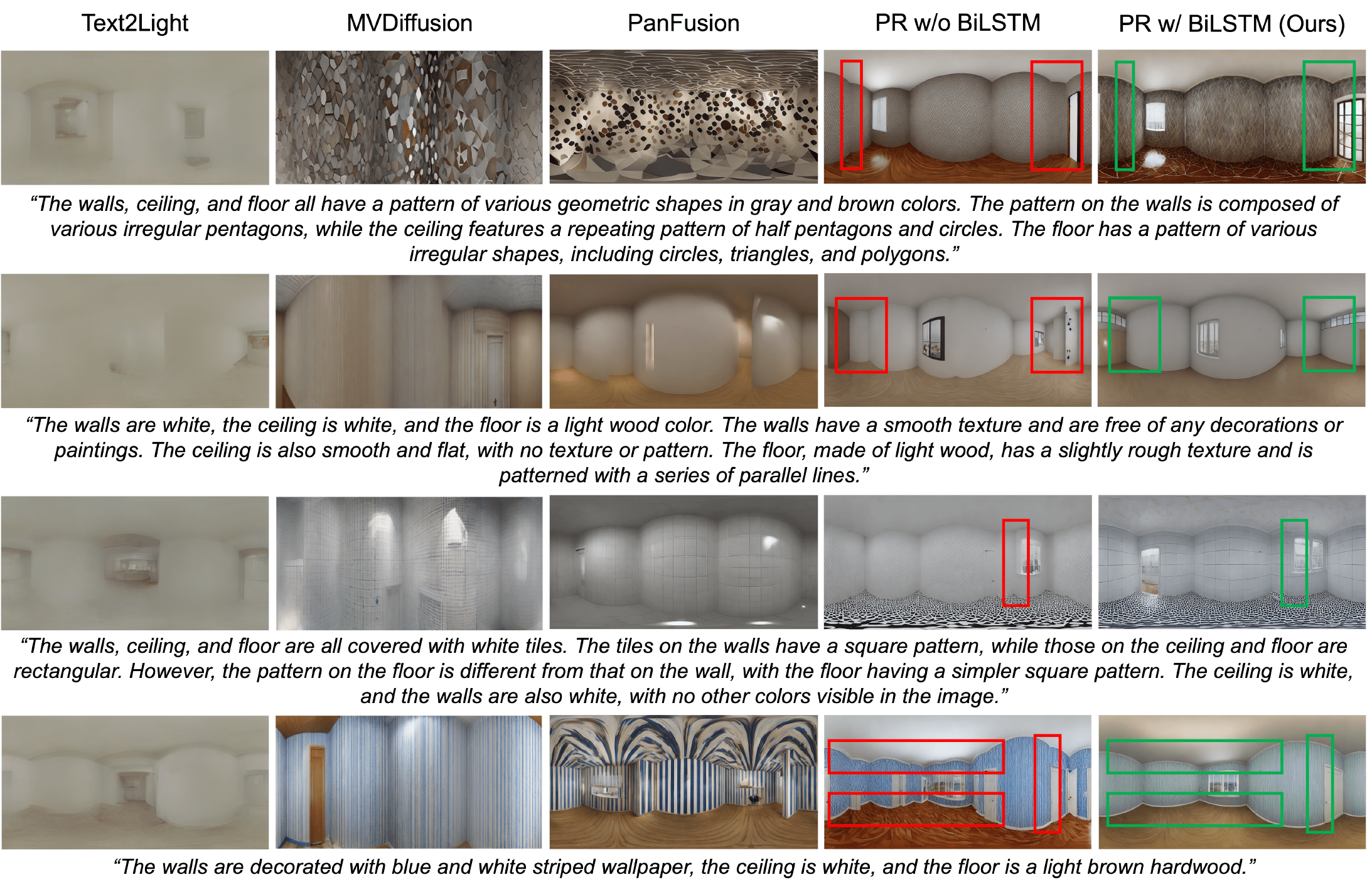}
\caption{\textbf{Qualitative comparisons on panorama image generation.} Red boxes denote wrong layout of the room, whereas green boxes denote correct layout of the room. PR indicates our Programmable-Room.}
\label{fig:exp_pano}
\end{figure}

\subsection{Comparison with State-of-the-arts}

\noindent\textbf{Panorama Image Generation.} We compare PRIG with other panorama image generation models \cite{chen2022text2light, tang2023mvdiffusion, zhang2024taming}. For a fair comparison, we trained each baseline with the same train dataset of Structure3D and its matching image captions which we used to train PRIG. We measured FID and KID by feeding the captions of the test split of Structure3D to each baseline as the input text prompt. 
As PRIG requires visual prompts such as a layout map, a depth map, and a semantic map, we generate the visual prompts within our framework by inserting the given text prompt to Programmable-Room. The quantitative results in Table. \ref{tab:exp_pano} prove the superiority of PRIG in generating panorama texture images. Our method achieves the best FID and KID scores.  
The comparably high scores of our method implies that the baselines have difficulties in displaying structural coherence and in reflecting the texture instructions. Specifically, in the result images of Text2Light and MVDiffusion in Fig. \ref{fig:exp_pano}, the left and the right ends are not continuous. Moreover, results generated by the baselines including PanFusion do not match the texture descriptions.

\begin{figure*}[t!]
\centering
\includegraphics[width=\textwidth]{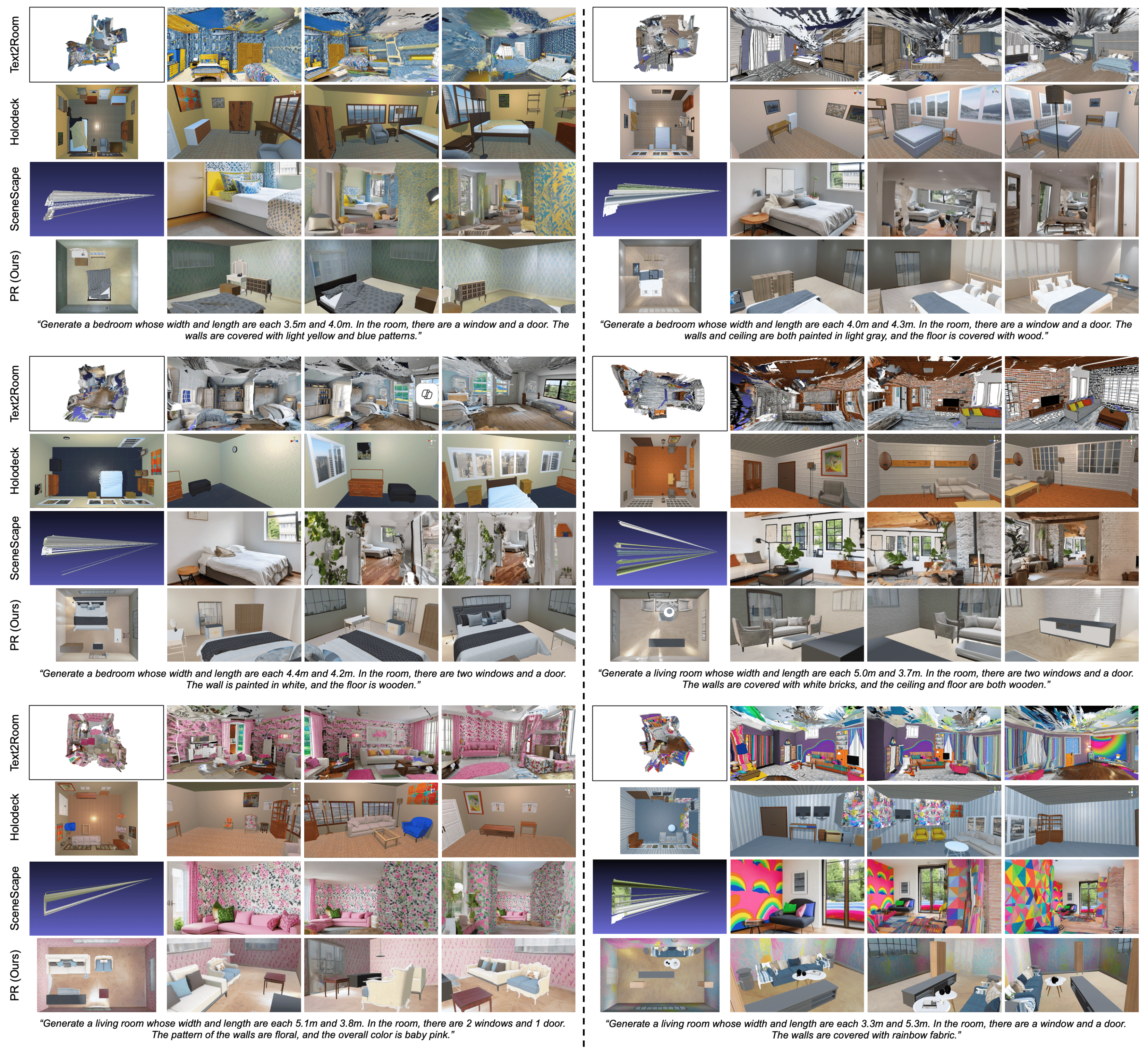}
\caption{\textbf{Qualitative comparisons on 3D mesh generation.} The first and fifth columns include rendered scenes from the top-view, whereas the rest columns include rendered scenes from the various views. PR indicates our Programmable-Room.}
\label{fig:exp_mesh}
\end{figure*}

\noindent\textbf{3D Mesh Generation.} We compare our method with Text2Room \cite{hollein2023text2room}, Holodeck \cite{yang2024holodeck}, and SceneScape \cite{fridman2024scenescape}. As shown in Table \ref{tab:exp_mesh}, our method achieves the highest scores both in PQ and 3DS. In terms of the layout, renderings from Text2Room and SceneScape demonstrate unrealistic shapes which resulted in low 3DS scores. Holodeck fails to reflect the specified room shapes, whereas our method generates a room which matches the target shape. In terms of the rendered image quality, our method better satisfies the given texture descriptions, achieving a higher PQ score. On the contrary, results from the baselines either fail to reflect the texture description or are visually unrealistic. For example, in the second case of Fig. \ref{fig:exp_mesh}, even though the instruction specifies the walls to be in light gray, the output of Text2Room is painted in purple and blue, and the output of Holodeck is painted in light orange. Moreover, the rooms generated by Text2Room contains floating artifacts, which degrades the perceptual quality. Lastly, rendered images from SceneScape are unrealistic for an indoor scene because it originally aims to generate a walk-through video.

\subsection{Ablation Studies}

\noindent\textbf{BiLSTM loss in PRIG.} Quantitative results in Table. \ref{tab:exp_pano} prove the effectiveness of implementing BiLSTM in our method, as both the FID and KID score decrease by applying BiLSTM. Moreover, as illustrated in Fig. \ref{fig:exp_pano}, PRIG without BiLSTM generates panorama images with inappropriate curves in the layout which are depicted in red boxes. However, PRIG with BiLSTM creates panorama images with more plausible layouts depicted in green boxes. It implies that 1D representation obtained from BiLSTM provides a long-range geometric pattern of room layout, helping PRIG to better reflect the given room layout.

\begin{figure}[t!]
\centering
\includegraphics[width=\linewidth]{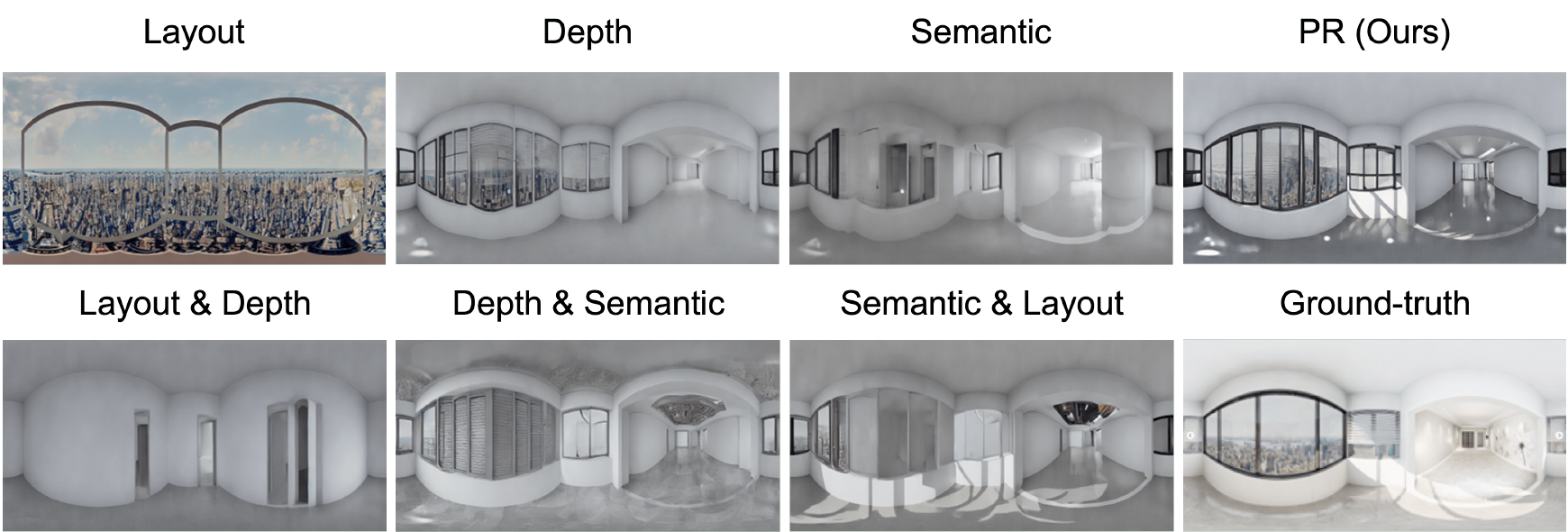}
\caption{\textbf{Qualitative results on types of visual prompts.} The result from Programmable-Room (PR) indicates all the three visual prompts are given.}
\label{fig:exp_prompt}
\end{figure}

\noindent\textbf{Types of Visual Prompts.} Quantitative results of PRIG, in Table \ref{tab:exp_prompt}, demonstrate the best performance when the combination of layout, depth and semantic maps are simultaneously given. As illustrated in Fig. \ref{fig:exp_prompt}, giving only the layout as the visual prompt to PRIG matches the room layout but cannot differentiate the floor, walls, and ceiling nor display appropriate depth for an indoor scene. When only the depth map is given, PRIG fails to correctly demonstrate the layout. On the other hand, conditioning only on the semantic map fails to reflect the geometry of the room. When two visual prompts are given, the results are better but still have artifacts. However, when three visual prompts are given, PRIG generates a reasonable output which has the faithful room shape and texture similar to the ground-truth image.

\begin{figure}[t!]
\centering
\includegraphics[width=\linewidth]{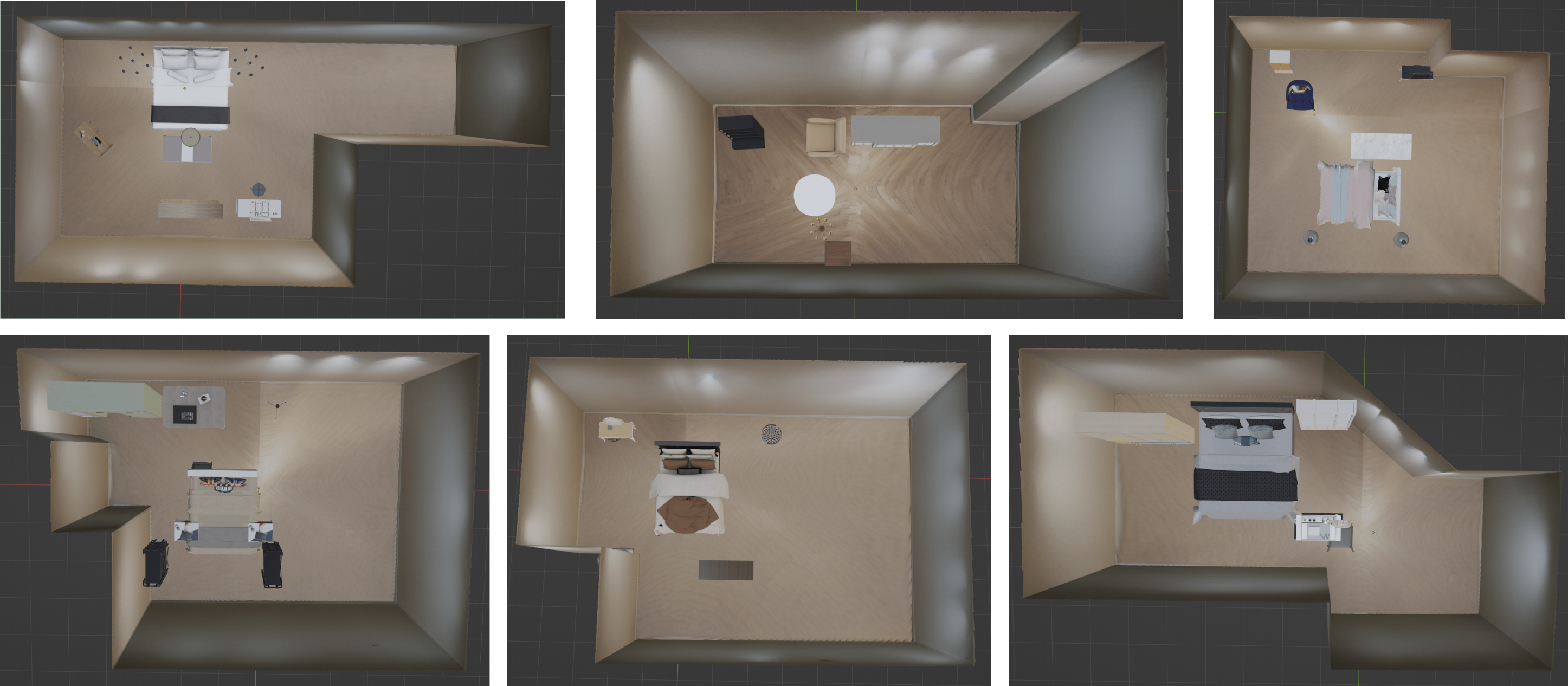}
\caption{\textbf{Qualitative results on the controllability of room layouts.} The top-view results demonstrate Programmable-Room's capability in controlling the room layouts.}
\label{fig:exp_ctrl}
\end{figure}

\noindent\textbf{Controllability of Room Layouts.} 
As shown in Fig. \ref{fig:exp_ctrl}, Programmable-Room is capable of generating rooms with various and complex layouts.
For user convenience, we provide choices to user to either specify the room shape with texts or use existing floor templates. 
Since PRIG is trained with multiple visual prompts conveying room's geometric information, user can robustly generate rooms with intricate layouts.

\begin{figure}[t!]
\centering
\includegraphics[width=\linewidth]{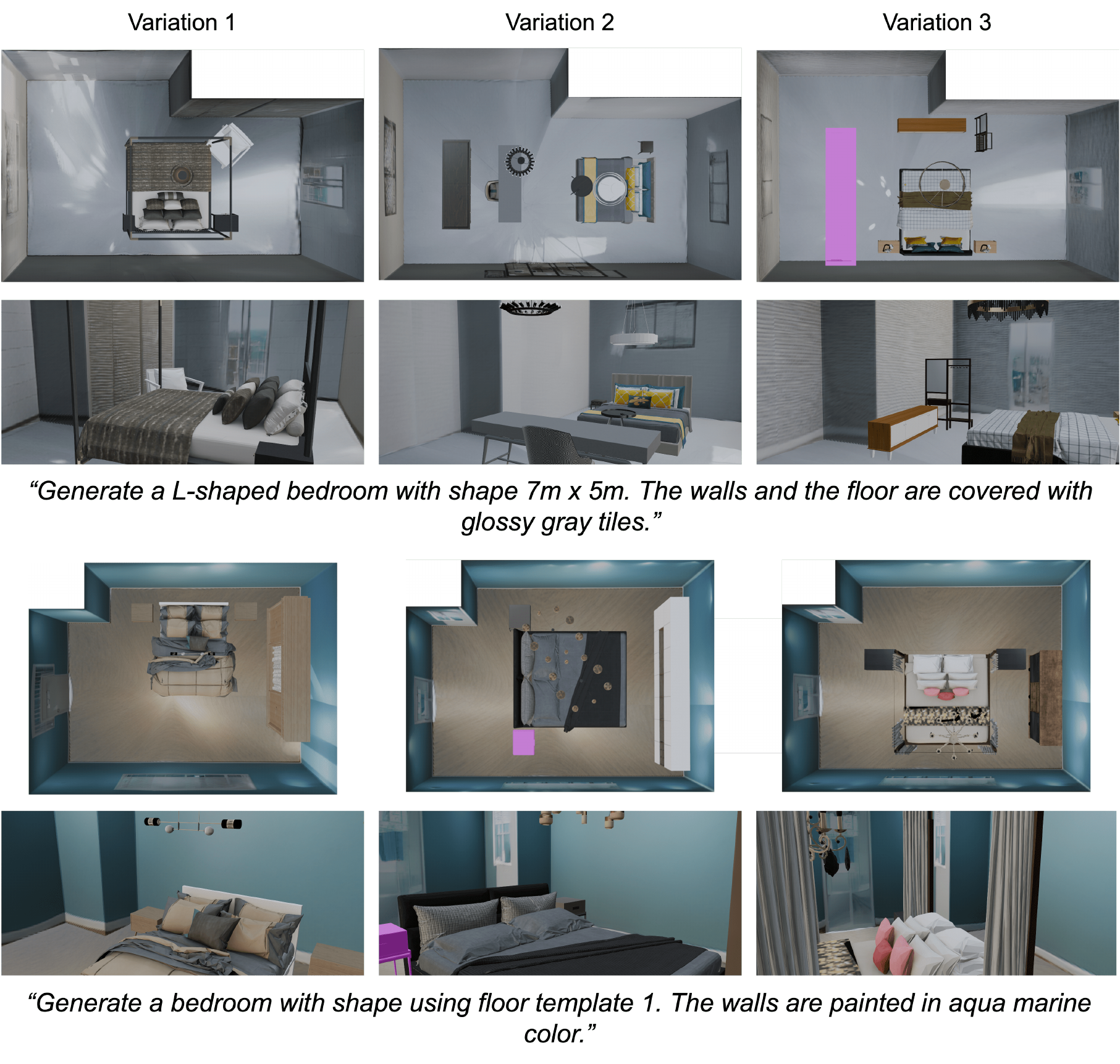}
\caption{\textbf{Qualitative results on diversity under same instructions.} Programmable-Room is capable of generating rooms with various textures and furniture layouts which still satisfy the user instructions.}
\label{fig:exp_diverse}
\end{figure}

\noindent\textbf{Diversity under Same Instructions.} 
As shown in Fig. \ref{fig:exp_diverse}, PRIG can understand a complicated instruction about the room’s shape such as \textit{“Generated a L-shaped bedroom”}, or users can even select predefined floor templates to generate the desired room shape. Moreover, from just one instruction, users can generate diverse rooms which still satisfy the user’s demands. 

\begin{figure}
\centering
\includegraphics[width=\linewidth]{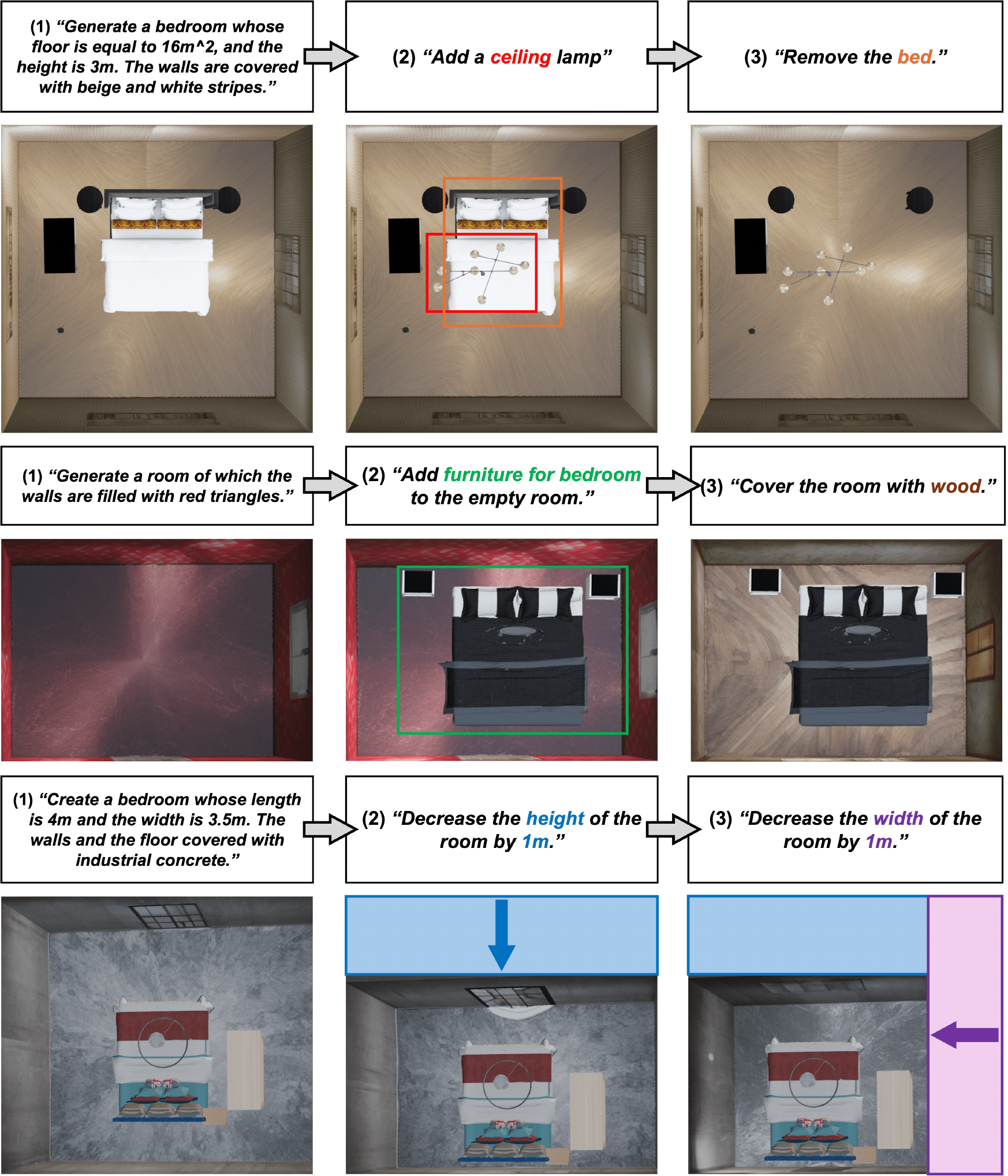}
\caption{\textbf{Additional editing results.} Users can continuously edit the room by providing additional instructions.}
\label{fig:exp_editing}
\end{figure}

\noindent\textbf{Additional Editing Results.}
We demonstrate additional editing results in Fig. \ref{fig:exp_editing}. Currently, with Programmable-Room, users can edit the furniture by adding, replacing or removing them, or change the texture or the shape of the generated rooms.

\begin{table}[t!]
\centering
\caption{Quantitative results on types of visual prompts. The performance improves when all visual prompts are used.}
\resizebox{0.7\linewidth}{!}{
\begin{tabular}{ccc|cc}
\Xhline{2pt}
{Layout} & {Depth} & {Semantic} & {FID$\downarrow$} & {KID$\downarrow$} \\
\hline
\cmark & & & 125.73 & 0.23876 \\
 & \cmark &  & 79.45 & 0.07734 \\
 &  & \cmark & 79.35 & 0.06557 \\
\cmark & \cmark &  & 76.90 & 0.03867 \\
 & \cmark & \cmark & 68.35 & 0.03207 \\
\cmark &  & \cmark & 75.59 & 0.04540 \\
\cmark & \cmark & \cmark & \textbf{65.68} & \textbf{0.02354} \\
\Xhline{2pt}
\end{tabular}
}
\label{tab:exp_prompt}
\end{table}

\section{Conclusion}

We present Programmable-Room, a framework for interactive 3D room mesh generation and editing using user-provided instructions in natural language format. To carry out the various subtasks in 3D room mesh generation and editing with a single unified framework, Programmable-Room implements visual programming, which grants flexibility and precise control over each attribute of the 3D room mesh. In addition, we design a novel diffusion-based model, panorama room image generation, which generates a panorama room image from text and multiple visual prompts. We verify that our Programmable-Room's flexibility in terms of generation and editing room meshes quantitatively and qualitatively.
The limitation of Programmable-Room lies in the limited room categories, primarily bedrooms and living rooms, due to its dependence on the current furniture generation models. We expect more realistic and diverse outputs and will continuously update modules to incorporate improvements in related fields.

\newpage
\section{Appendix}

For Appendix, we discuss about how to obtain visual prompts during inference time of our Programmable-Room.

\subsection{Obtaining Visual Prompts During Inference}
In Programmable-Room, \textit{GenLayout} generates a layout map $L$ through an equirectangular projection of 3D coordinates of corners and lines connecting the corners of the target room mesh. This involves converting Cartesian coordinates to spherical coordinates:
\begin{equation}
(r, \theta, \phi)=Project_{spherical}(x, y, z),
\end{equation}
where $x$, $y$, and $z$ denote the coordinate along the x-axis, y-axis, and z-axis, respectively. In addition, $r = \sqrt{x^{2} + y^{2} + z^{2}}$, $\theta = \arctan\left(\frac{x}{y}\right)
$, and $\phi = \arccos\left(\frac{r}{z}\right)$. Then, the spherical coordinates are mapped to UV space coordinates as follows:
\begin{equation}
(u, v)=Project_{UV}(r, \theta, \phi),
\end{equation}
where $u = \frac{\theta}{2\pi}$ and $v = \frac{\phi}{\pi}$.

To obtain a depth map $D$, we use mathematical equations instead of off-the-shelf depth estimators, as only locations of the corners of the room are given. Thus, the depth of floor, ceiling, and walls are calculated separately using \textit{GenDepth}. First, the per-pixel vertical angle $a$ in radiance is calculated as follows:
\begin{equation}
a = \left(\frac{y}{h}\right) \pi,
\end{equation}
where $h$ is the height of the target panorama image. Then the depth of floor, ceiling, and walls are calculated using the following formulas:
\begin{equation}
D_{floor} = \left| \frac{y_{floor}}{\sin(a)} \right|,
\end{equation}
\begin{equation}
D_{ceiling} = \left| \frac{y_{ceiling}}{\sin(a)} \right|,
\end{equation}
\begin{equation}
D_{wall} = \left| \frac{cs}{\cos(a)} \right|,
\end{equation}
where $cs$ denotes the wall to camera distance on the horizontal plane at cross camera center. Finally, we obtain $D$ from \textit{GenDepth} by applying semantic masks for each floor, ceiling, and wall to the depth values. Lastly, we obtain a semantic map $M$ from \textit{GenSemantic} by converting $L$ into a binary image, detecting contours, and classifying each segment into ceiling, wall, and floor labels based on the center of $y$ value. Then, morphological closing is applied to fill in the gaps among the contours.
Therefore, we finally obtain visual prompts $L$, $D$, and $M$ during inference time.

\bibliographystyle{IEEEtran}
% \bibliography{main}

\newpage

\begin{IEEEbiography}[{\includegraphics[width=1in,height=1.25in,clip,keepaspectratio]{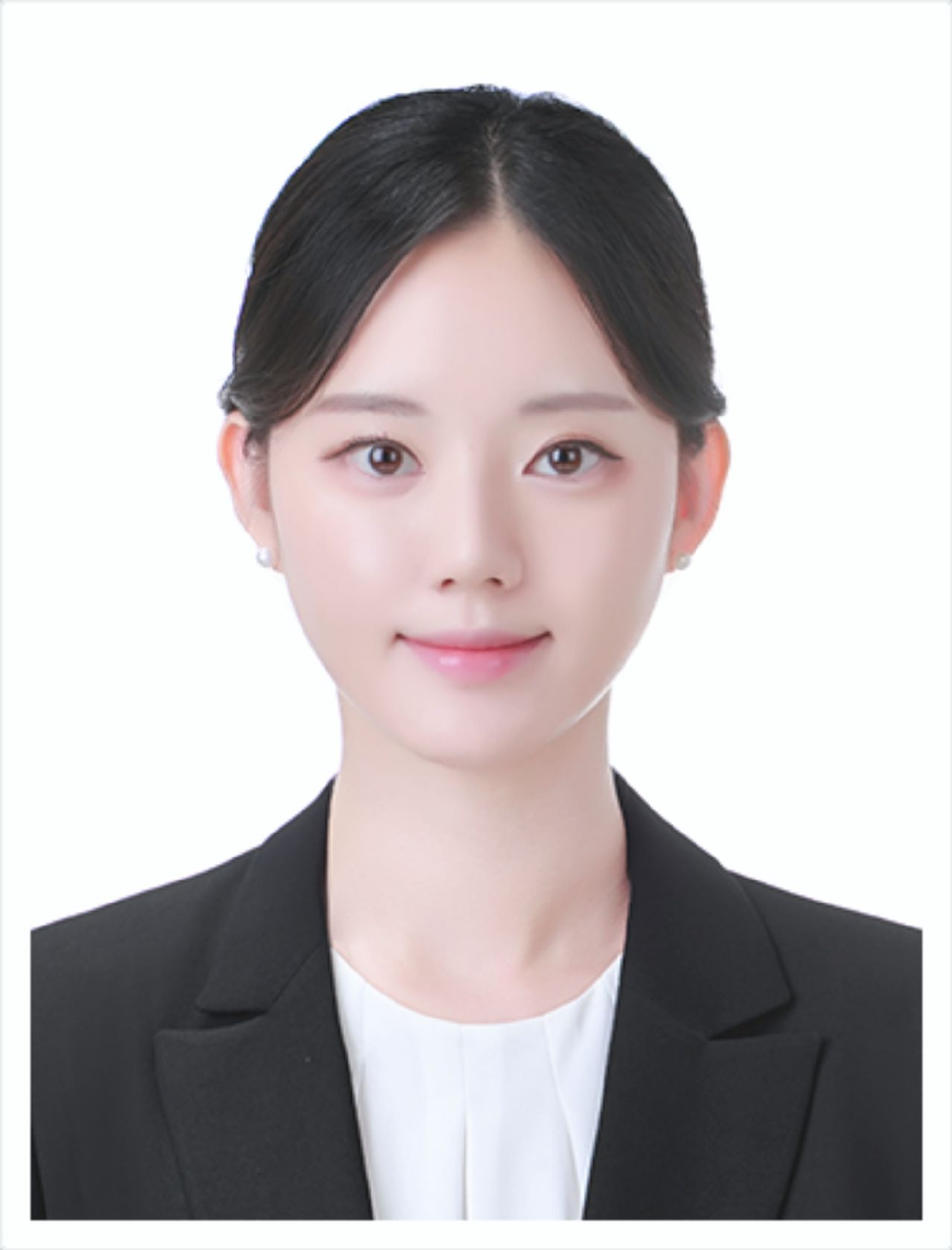}}]{\textbf{Jihyun Kim}} received the B.S. degree in business management from Sogang University, Seoul, South Korea, in 2021, and the M.S. degree in artifical intelligence from Sogang University, Seoul, South Korea, in 2024. Her current research interests include computer vision and deep learning.
\end{IEEEbiography}

\begin{IEEEbiography}[{\includegraphics[width=1in,height=1.25in,clip,keepaspectratio]{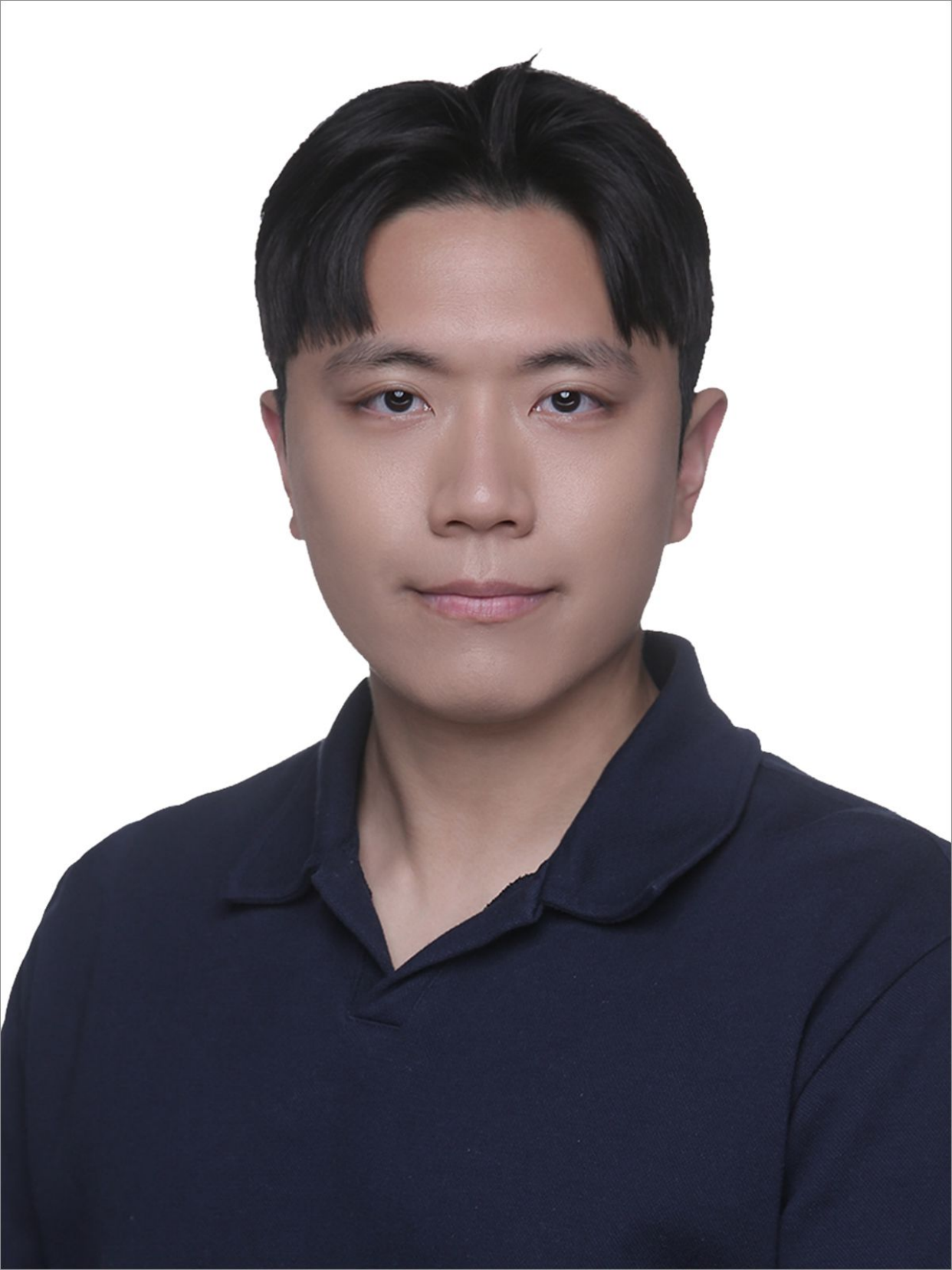}}]{\textbf{Junho Park}} received the B.S. degree in mathematics and electronics engineering (double major) from Sogang University, Seoul, South Korea, in 2022, and the M.S. degree in electrical engineering from Sogang University, Seoul, South Korea, in 2024. His current research interests include computer vision and deep learning.
\end{IEEEbiography}

\begin{IEEEbiography}[{\includegraphics[width=1in,height=1.25in,clip,keepaspectratio]{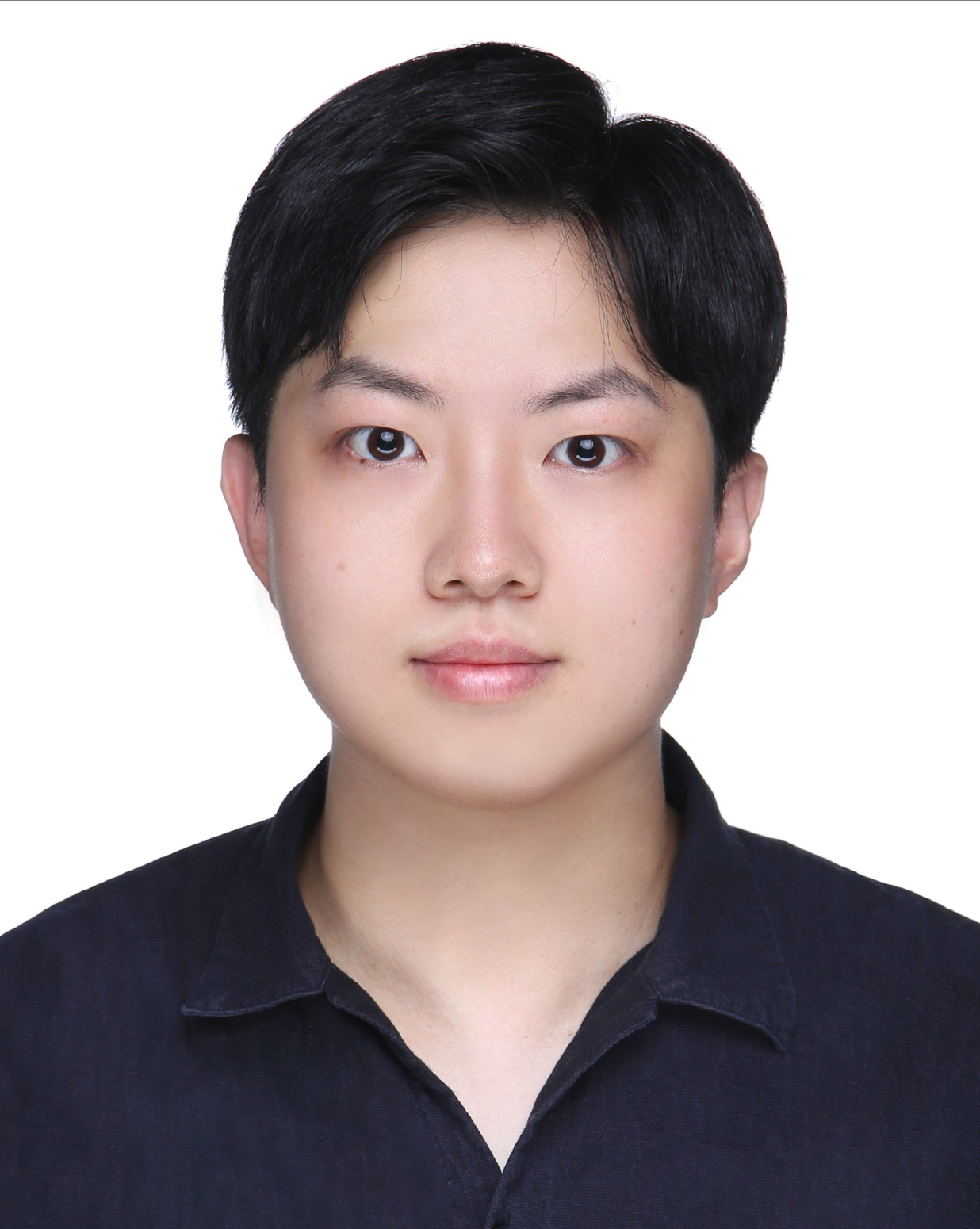}}]{Kyeongbo Kong}
received the B.S. degree in electronics engineering from Sogang University, Seoul, South Korea, in 2015, and the M.S. and Ph.D. degrees in electrical engineering from the Pohang University of Science and Technology (POSTECH), Pohang, South Korea, in 2017 and 2020, respectively. From 2020 to 2021, he was worked as a Postdoctoral Fellow with the Department of Electrical Engineering, POSTECH, Pohang, South Korea. 
From 2021 to 2023, he was an Assistant Professor of Media School at Pukyong National University, Busan. He is currently an Assistant Professor of Electrical and Electronics Engineering at Pusan National University.
His current research interests include image processing, computer vision, machine learning, and deep learning.
\end{IEEEbiography}

\begin{IEEEbiography}[{\includegraphics[width=1in,height=1.25in,clip,keepaspectratio]{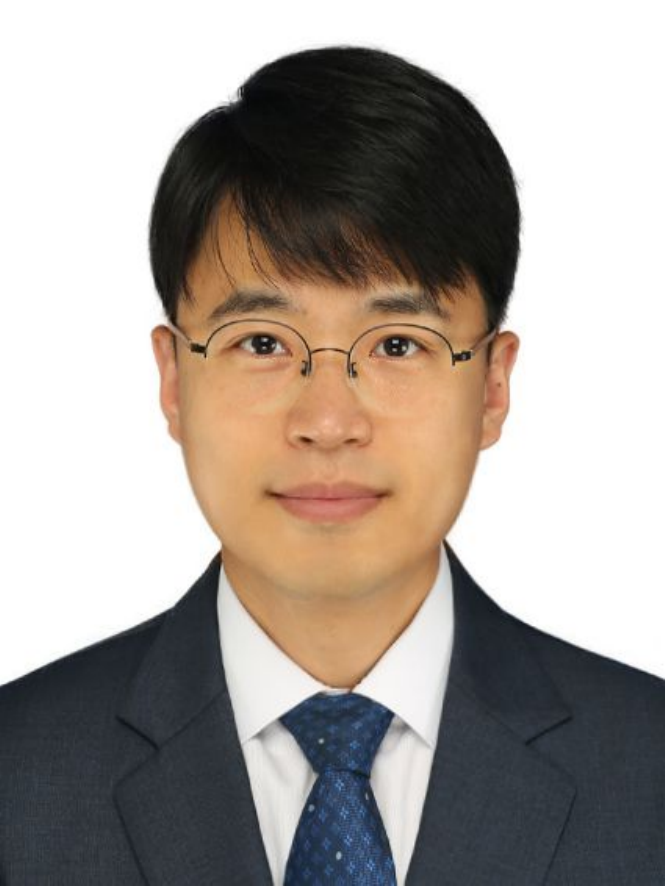}}]{\textbf{Suk-Ju Kang}}
(Member, IEEE) received the B.S. degree in electronic engineering from Sogang University, Seoul, South Korea, in 2006, and the Ph.D. degree in electrical and computer engineering from the Pohang University of Science
and Technology, Pohang, South Korea, in 2011. From 2011 to 2012, he was a Senior Researcher with LG Display Co., Ltd., Seoul, where he was a Project Leader for resolution enhancement and multiview 3-D system projects. From 2012 to 2015, he was an Assistant Professor of Electrical Engineering with Dong-A University, Busan, South Korea. He is currently a Professor of Electronic Engineering with Sogang University, Seoul. His current research interests include image analysis and enhancement, video processing, multimedia signal processing, digital system design, and deep learning systems. Dr. Kang was a recipient of the IEIE/IEEE Joint Award for Young IT
Engineer of the Year in 2019 and the Merck Young Scientist Award in
2022.
\end{IEEEbiography}

\end{document}